\documentclass[10pt,twocolumn,letterpaper]{article}
\usepackage{cvpr}

\usepackage{times}
\usepackage{epsfig}
\usepackage{graphicx}
\usepackage{amsmath}
\usepackage{amssymb}

\usepackage[pagebackref=true,breaklinks=true,colorlinks,bookmarks=false]{hyperref}

\usepackage{amssymb} 

\usepackage{enumitem}
\usepackage{booktabs}

\usepackage{amsmath}
\usepackage{xspace}

\usepackage[dvipsnames,table]{xcolor}
\usepackage[pagebackref=true,breaklinks=true,colorlinks,bookmarks=false]{hyperref}

\usepackage{cleveref}
\definecolor{darkergreen}{RGB}{21, 152, 56}
\definecolor{red2}{RGB}{252, 54, 65}

\definecolor{lightblue}{rgb}{0,0,1}

\usepackage{arydshln}

\usepackage{pifont}
\usepackage[table]{xcolor}
\usepackage{algorithm}
\usepackage{algorithmic}

\newcommand*\colorcmark[1]{%
  \expandafter\newcommand\csname #1cmark\endcsname{\textcolor{#1}{\ding{51}}}%
}
\colorcmark{green}

\newcommand*\colorxmark[1]{%
  \expandafter\newcommand\csname #1xmark\endcsname{\textcolor{#1}{\ding{55}}}%
}
\colorxmark{red}

\usepackage{multirow}
\usepackage{enumitem}
\usepackage{subcaption}
\usepackage[table]{xcolor}
\usepackage{amsthm}
\newlength\savewidth\newcommand\shline{\noalign{\global\savewidth\arrayrulewidth
  \global\arrayrulewidth 1pt}\hline\noalign{\global\arrayrulewidth\savewidth}}

    \newlength\thinwidth
    \definecolor{Gray}{gray}{0.92}
    \newcolumntype{a}{>{\columncolor{Gray}}c}
    \definecolor{LightCyan}{rgb}{0.88,1,1}
    \definecolor{lightblue}{rgb}{0,0,1}
    \definecolor{darkergreen}{RGB}{21, 152, 56}
    \definecolor{highlightRowColor}{gray}{0.92}

\crefname{section}{Sec.}{Secs.}
\Crefname{section}{Section}{Sections}
\Crefname{table}{Table}{Tables}
\crefname{table}{Tab.}{Tabs.}

\usepackage{bm}

\usepackage{pifont}
\usepackage{minitoc}
\usepackage[table]{xcolor}
\definecolor{darkergreen}{RGB}{21, 152, 56}
\definecolor{red2}{RGB}{252, 54, 65}

\definecolor{maroon}{cmyk}{0,0.87,0.68,0.32}
\definecolor{aliceblue}{rgb}{0.94, 0.97, 1.0}

\definecolor{citecolor}{HTML}{0071BC}
\definecolor{linkcolor}{HTML}{ED1C24}

\definecolor{LightCyan}{rgb}{0.92,1,1}
\definecolor{Gray}{gray}{0.9}

\definecolor{lightRed}{rgb}{0.8,0,0}
\definecolor{lightGreen}{rgb}{0,0.8,0}

\definecolor{deemph}{gray}{0.6}
\newcommand{\gc}[1]{\textcolor{deemph}{#1}}

\begin{document}

\title{Object-aware Video-language Pre-training for Retrieval}

\author{Alex Jinpeng Wang$^{1}$ \quad Yixiao Ge$^{2}\;$\quad Guanyu Cai$^{1,5}$ \quad Rui Yan$^1$ \quad Xudong Lin$^4$ \\ Ying Shan$^2$ \quad Xiaohu Qie$^{3}$  \quad Mike Zheng Shou$^1$\thanks{Corresponding Author.}\\ \\
$^1$Show Lab, National University of Singapore\quad $^2$ARC Lab,$^3$Tencent PCG\quad \\ $^4$Columbia University\quad $^5$Tongji University\\}
\maketitle

\begin{abstract}

Recently, by introducing large-scale dataset and strong transformer network, video-language pre-training has shown great success especially for retrieval. 
Yet, existing video-language transformer models do not explicitly explore fine-grained semantic alignment.
In this work, we present Object-aware Transformers, an object-centric approach that extends video-language transformer to incorporate object representations.
The key idea is to leverage the bounding boxes and object tags to guide the training process.
We evaluate our model on three standard sub-tasks of video-text matching on four widely used benchmarks. 
We also provide deep analysis and detailed ablation about the proposed method.
We show clear improvement in performance across all tasks and datasets considered,
demonstrating the value of a model that incorporates object representations into a
video-language architecture.
The code will be released at \url{https://github.com/FingerRec/OA-Transformer}.
   
\end{abstract}

\section{Introduction}
\label{sec:intro}

Learning scalable video-text representations for retrieval requires the understanding of both visual and textual clues, as well as the semantic alignment between these two modalities. 
Large-scale contrastive-based pre-training methods \cite{bain2021frozen,lei2021less} dominate the recent literature, where a ``dual-encoder'' framework (a video encoder and a text encoder) is trained 
in an end-to-end manner.
Although these methods have led to great performance advances,
\textit{we figure out that the lack of regularization on fine-grained semantic associations hinders their further improvements}.


\begin{figure}[t]
  \centering
\includegraphics[width=.85\linewidth]{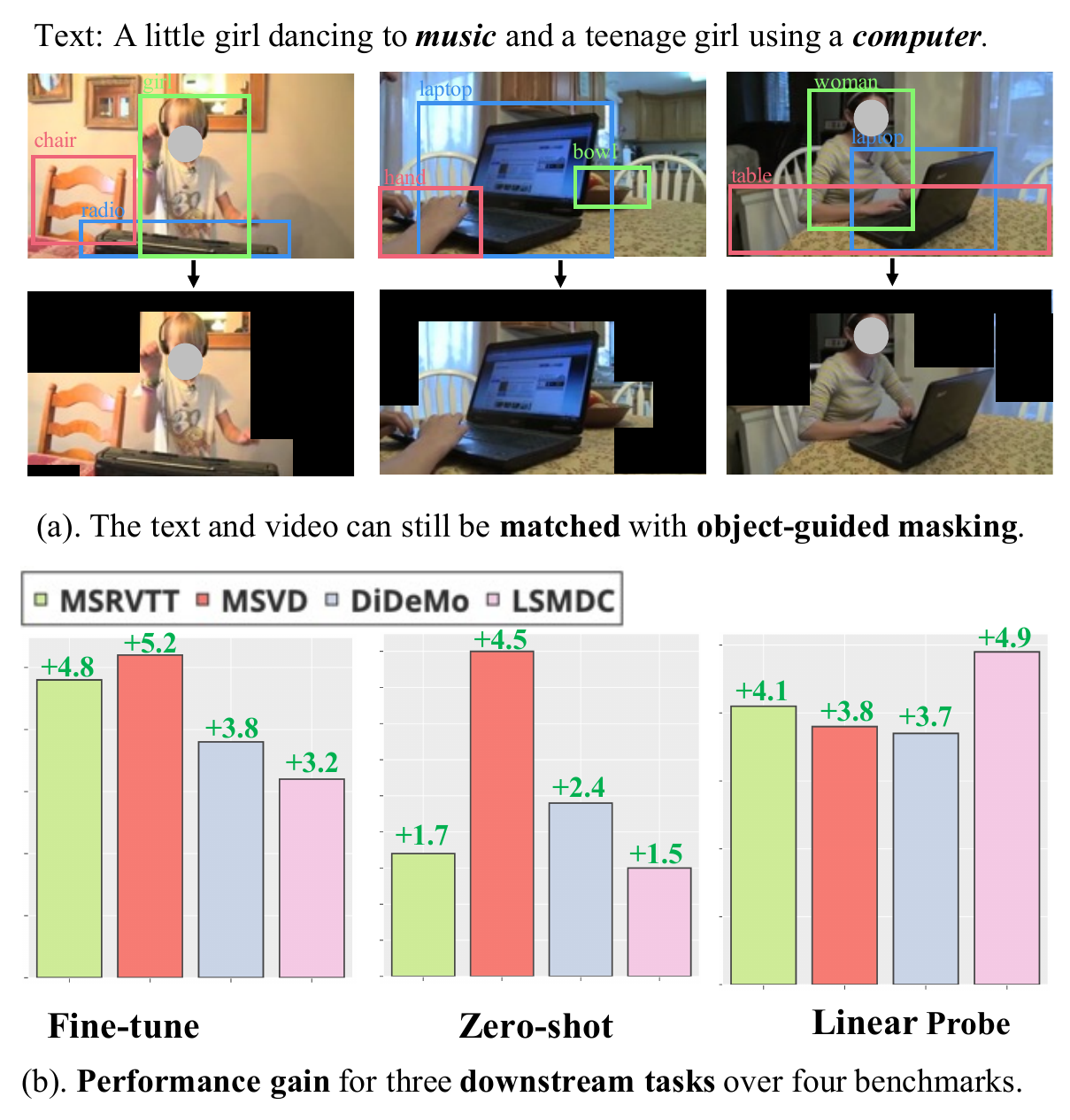}
   \caption{
   \textbf{(a). Masking object-irrelevant region keep the semantic unchanged.}
   From this example, we observe:
   1. The object region is highly overlapped with visual salient region.
   2. The predicted \textcolor{lightGreen}{Object Tags} has semantic relation with caption. \emph{e.g.}, \textit{Music} and \textit{Ratio}.
   \textit{Laptop} and \textit{Computer}.
   \textbf{(b). Our method vs. SOTA on three downstream tasks.}
   Motivated by (a), by incorporating the object into the learning of video-language pretraining with simple \textcolor{lightGreen}{Object-guided Masking}, we show promising results over multiple downstream video-language tasks.
   }
   \label{fig:1_object_motivation}
\end{figure}

Thanks to the great progress of image-text pre-training \cite{li2020oscar,chen2020uniter,lu2019vilbert,Su2020VL-BERT:,li2019visualbert,tan2019lxmert,zhang2021vinvl},
a series of methods attempt to leverage an off-the-shelf object detection model to generate richer information for cross-modality understanding, including the visual objects and their tag concepts.
The object information, together with the raw image and sentence, are then fed into a joint encoder for cross-modality interaction, leading to better correlations between regions and phrases.
Given the success of object information in image-text pre-training,
\textit{
it is intuitive to exploit the objects to improve video-text retrieval.
}
However, there exist some main challenges that prevent us from na\"ively employing existing object-based techniques on video-text pre-training.

Fig.~\ref{fig:1_object_motivation}(a) shows that \textit{object boxes and tags always focus on the salient regions and semantics}, which are considered as the most important in each video. 
Existing object-based image-text pre-training methods either adopt an image-text joint encoder \cite{li2020oscar,li2019visualbert} or cross-modality co-attention modules \cite{lu2019vilbert} for interaction between cross-modality local features. 
Despite the results being positive, it is impractical to adapt this paradigm from image domain to video domain. This is because all these methods require pre-extracted offline object feature for whole dataset.
It would lead to \textbf{unaffordable computational overhead} to extract all objects, due to the billion-level frames.
Moreover, their downstream performance heavily depends on the quality of the objects since they also need the objects as input for inference.

\begin{figure}[t]
  \centering
\includegraphics[width=.9\linewidth]{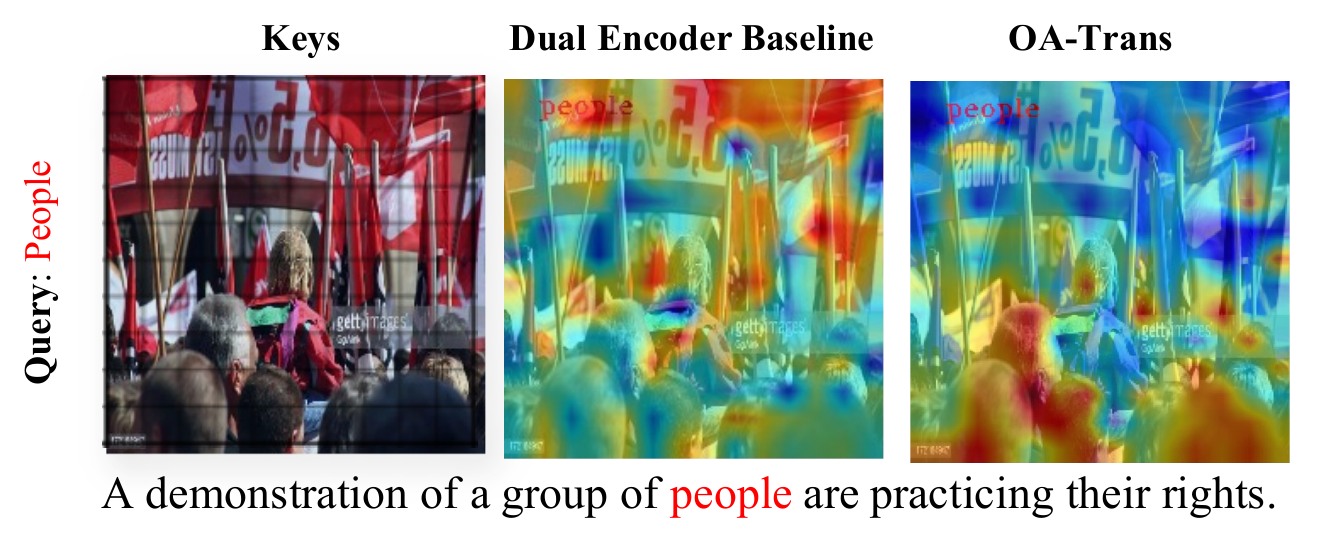}
   \caption{
   \textbf{Visualization of the cross-modality attention on a video-text sample.}
   This video is retrieved by the baseline dual encoder network \cite{bain2021frozen} \textcolor{lightRed}{wrongly} but \textcolor{lightGreen}{correctly} by our Object-aware Transformer (OA-Trans).
   }
   \label{fig:1_dual_align}
\end{figure}

To this end,
we introduce a simple yet effective paradigm for video-text pre-training, namely \textbf{Object-aware Transformer (OA-Trans)}, \textit{which explicitly enhances the fine-grained video-text interaction of the dominant ``dual-encoder'' framework at the same time maintaining its retrieval efficiency during inference.}
This is achieved by two novel designs in our method as follows.

(1) \textbf{Single anchor frame that encodes object information}.
Instead of replacing all sampled video frames with their extracted object regions, we balance the matching recall and efficiency via combining whole frames together with a novel anchor frame that encodes object information. Specifically, we propose to only extract object regions on this anchor frame and softly mask out the non-object regions on this anchor frame. 

(2) \textbf{A novel 4-stream object-aware contrastive (OAC) loss}.
The input to our OA-Trans for pretraining include four stream: \textit{raw video}, \textit{anchor frame}, \textit{object tags} (predicted object categories), and \textit{raw text}.
To explore how to combine these four streams, we do extensive experimental explorations and find out it works the best to contrast the \textit{raw video} stream with the \textit{object tags} stream and the \textit{raw text} stream with the \textit{anchor frame} stream.
Note that the objects are only used for pre-training in our method, so the quality of detection has less effect on the downstream tasks and we do not need any extra computational overhead for downstream retrieval.
As shown in Figure \ref{fig:1_dual_align}, a dual-network spreads its attention over the whole frame randomly while OA-Trans with OAC loss can successfully focus on the ``People'' region.

Our contributions are as follows:
\begin{itemize}
\item We are the first to successfully develop an object-aware dual encoder model, namely OA-Trans, for end-to-end video-language pre-training.
\item To alleviate the heavy cost of extracting object boxes, we propose to unify sampled whole frames with a single anchor frame whose non-object regions have been masked.
\item We design a novel object-aware contrastive loss based on our unique input streams of video frames, textual query, the masked image, and predicted object tags on the anchor object frame.
\item Our OA-Trans achieves significant improvements of Recall@1 on 4 benchmarks with three downstream tasks (Figure~\ref{fig:1_object_motivation} (b)).
\emph{e.g.} MSVD (from 46.2\% to 51.4\%).
\end{itemize}

\section{Related Work}

\subsection{Video-Language Pretraining}
Limited by small-scale video-language datasets, previous video-language pretraining methods~\cite{wangdig,yu2018joint,liu2019use,miech18learning,gabeur2020multi}, have tended to use a combination of multiple ``experts" to extract multi-modal features offline, e.g., face, scene, object recognition action recognition, sound classification, and optical character recognition.

However, since a large-scale video-language dataset,  HowTo100M~\cite{miech2019howto100m}, was proposed, there has been a trend of leveraging pretraining on large-scale data to learn better video-language representations. Most of these video-language pretraining methods~\cite{luo2020univilm,patrick2020support,sun2019videobert,amrani2020noise,li2020hero} use a space-time CNN to pre-extract video features and propose a fusion module to align video features with language features that share the same semantics. Recently, considering most space-time CNNs are trained on Kinetics~\cite{Kinetics} that is much smaller than the pretraining dataset, to fully utilize massive information in pretraining datasets, end-to-end pretraining methods, ClipBert~\cite{lei2021less} and Frozen~\cite{bain2021frozen} are proposed. 

\begin{figure*}[h]
  \centering
   \includegraphics[width=.82\linewidth]{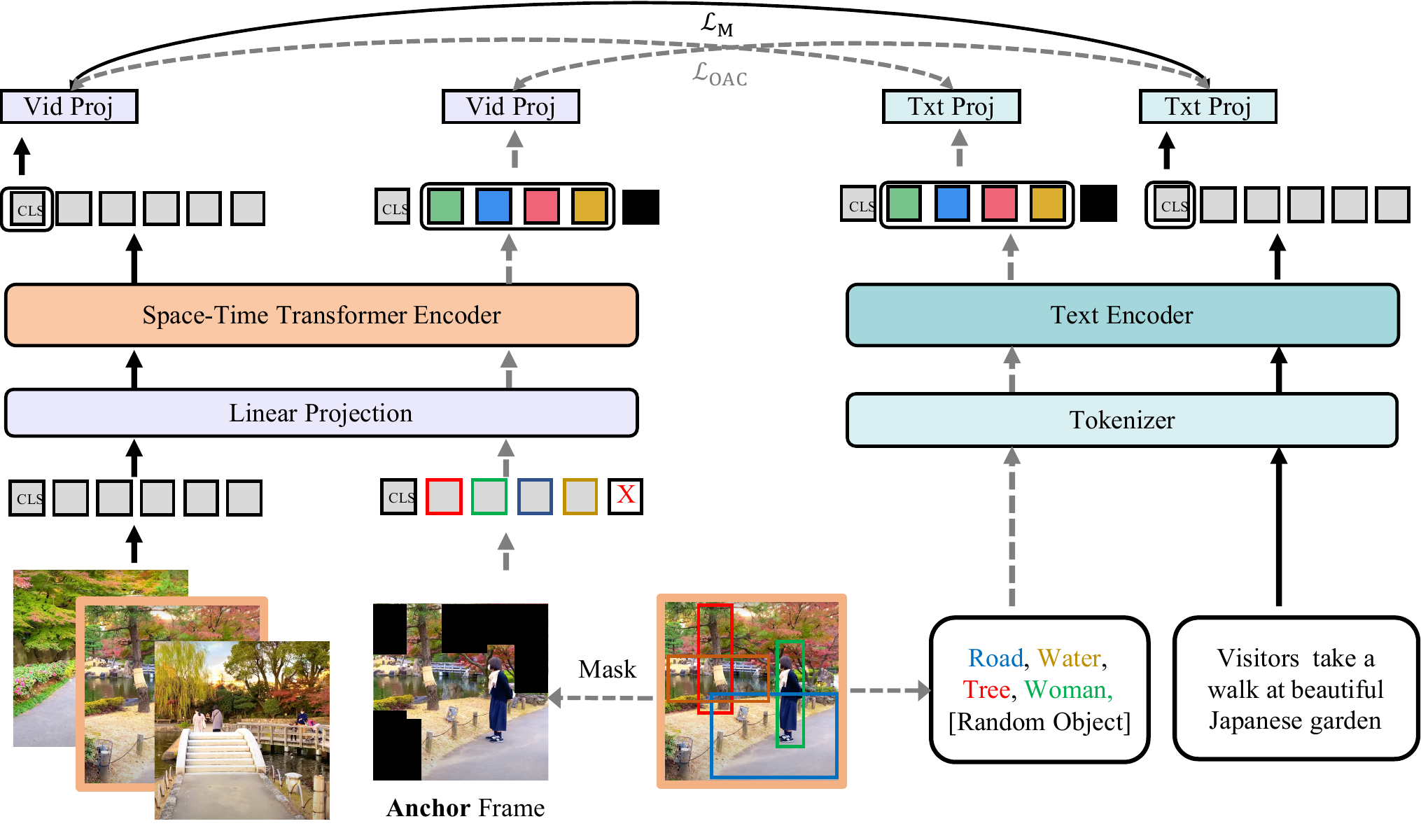}

   \caption{
   \textbf{
   Illustration of our Object-aware Transformer (OA-Trans).
   }
   The grey dotted line means the data flow are used only for pretrain but downstream tasks.
   The object tags and corresponding region guide the model to learn to attend to discriminative objects.
   } 
   \label{fig:3_ppl}
\end{figure*}

\subsection{Object in Vision-Language Tasks}
Recently, object-centric models have been successfully applied in various vision-language tasks, such as visual question answering~\cite{anderson2018bottom}, image captioning~\cite{anderson2018bottom}, image-text retrieval~\cite{faghri2017vse++,lee2018stacked} and image-text pretraining~\cite{li2020oscar,chen2020uniter,lu2019vilbert,Su2020VL-BERT:,li2019visualbert,tan2019lxmert}. Especially in the field of image-text pretraining, since the proposal of Bottom-Up Top-Down attention (BUTD)~\cite{anderson2018bottom}, fine-grained features extracted from the level of objects gradually becomes the most common inputs of image-text pretraining models~\cite{li2020oscar,chen2020uniter,lu2019vilbert,Su2020VL-BERT:,li2019visualbert,tan2019lxmert}. Benefited from the object features that are the salient image regions and can be easily aligned with textual features, object-centric image-text models ~\cite{li2020oscar,chen2020uniter,lu2019vilbert,Su2020VL-BERT:,li2019visualbert,tan2019lxmert} learn well-aligned image-text representations.

Although object-centric models have achieved remarkable progress in image-text pretraining, there lacks further exploration in video-text domain. ActBert~\cite{zhu2020actbert} leverages object features to achieve better language-and-visual alignment. 
However, it needs to extract object features over the whole video and cooperate with features from other feature encoders. The computation cost of extracting object features and domain gap between different feature encoders prevent ActBert from an efficient and powerful object-centric model. 
Thus, how to appropriately bring object-level features to video-language pretraining still remains unsolved.
In this work, to address the aforementioned issues, we propose an object-aware transformer to integrate object regions into video transformer~\cite{bertasius2021spacetime} seamlessly.

\section{Approach}

The human visual system tends to focus on objects and other salient image regions \cite{quinton1979objects, chao2000representation}.
Representing video semantics using objects facilitates compositional semantic understanding, because many perceptual components remain similar for a kind of object.
Thus, a model that captures this compositional aspect potentially pays less attention to semantic-irrelevant information.
Bringing this motivation into mind, we first revisit the current dual encoder framework in Section \ref{sec:revisit}, which our model extends, and present Object-aware Transformer (OA-Trans) in Section \ref{sec:oa_transformer}.
We further discuss the advantages of OA-Trans and different ways to utilize object information in Section \ref{sec:discussion}.


\subsection{Dual-encoder Framework}
\label{sec:revisit}
The earlier works in video-language pretraining focus on aligning the raw-pixel video and raw text with a contrastive loss in both dual-encoder framework \cite{bain2021frozen,lei2021less} and one stream \cite{zellers2021merlot} framework.
In this work, we choose the simple and effective dual-encoder framework (independent visual encoder and text encoder) Frozen \cite{bain2021frozen} as our baseline.
For the visual stream, a video project head is laid at the top of the visual encoder to project the output \emph{cls} embedding into a shared embedding space.
Similar to the visual stream, a text projection head is also laid at the top of the text encoder to project the \emph{cls} token of text into shared embedding space.
The same as text stream and the normalized embedding of video and text recorded as $v$ and $t$, respectively.

\paragraph{Objective:}
To train this dual-encoder framework, the normalized embedding of matched text-video pairs in the batch are treated as positives, and all other pairwise combinations in the batch are treated as negatives.
In practice, supposing we have $K$ samples in a batch, then the symmetrical contrastive loss is introduced as follows:
\begin{align}
    \mathcal{L}_{v2t} = - log\frac{\exp(sim(v,t))/\tau}{\sum_{i=0}^{K} \exp(sim(v^i,t))}
\end{align}
\begin{align}
        \mathcal{L}_{t2v} = - log\frac{\exp(sim(t,v))/\tau}{\sum_{i=0}^{K} \exp(sim(t^i,v))},
\end{align}
where $\tau$ is the temperature and $sim$ is a similarity function (\emph{i.e.}, dot product).
The final video-text matching loss is $\mathcal{L}_{{M}}=\mathcal{L}_{v2t} + \mathcal{L}_{t2v}$.

\begin{table*}[t]
\centering
\footnotesize
\begin{tabular}{@{}llllrrrr@{}}
\textbf{Method} & \textbf{Years} & \textbf{Vis Enc. Init.} &\textbf{Pretrained Data} & \textbf{R@1} & \textbf{R@5} & \textbf{R@10} & \textbf{MedR} \\ 
\shline
ActBERT~\cite{zhu2020actbert} &CVPR'20  & VisGenome &[136M] HowTo100M& 16.3 & 42.8 & 56.9 & 10.0  \\
VidTranslate~\cite{korbar2020video} &Arxiv'20& IG65M &[136M] HowTo100M & 14.7 & - & 52.8  \\
NE\cite{amrani2020noise} &AAAI'21& ImageNet, Kinetics  &[136M] HowTo100M  & 17.4 & 41.6 &  53.6 & 8.0  \\
ClipBERT~\cite{lei2021less} &ICCV'21 & - &[5.6M] COCO, VisGenome & 22.0 & 46.8 & 59.9 & 6.0   \\  MMT~\cite{gabeur2020multi}&ECCV'20 &  Numerous experts &[136M] HowTo100M  & 26.6 & 57.1 & 69.6 & 4.0 \\
Frozen \cite{bain2021frozen}  &ICCV'21& ImageNet & [3M] CC3M& 25.5  & 54.5  & 66.1  & 4.0 \\
Frozen \cite{bain2021frozen} &ICCV'21 & ImageNet  & [5.5M] CC3M, WebVid-2M & 31.0 & 59.5 & 70.5 & 3.0 \\
\rowcolor{LightCyan}
Frozen[Our Imp.] &ICCV'21 & ImageNet  & [5.5M] CC3M, WebVid-2M & 33.2 & 61.5& 71.9 & 3.0 \\
Support Set~\cite{patrick2020support}&ICLR'21 & IG65M, ImageNet&[136M] HowTo100M & 30.1 & 58.5 &69.3 & \textbf{3.0}  \\
\textbf{OA-Trans} & & ImageNet & [2.5M] Webvid-2M & \textbf{32.7}  & \textbf{60.9}  & \textbf{72.5}  & \textbf{3.0} \\ 
\textbf{OA-Trans}  && ImageNet  & [5.5M] CC3M, WebVid-2M  & \textbf{35.8} & \textbf{63.4} & \textbf{76.5} & \textbf{3.0} \\ 
\gc{\textbf{OA-Trans$\ddag$}} & &  \gc{CLIP-WIT}  & \gc{[5.5M] CC3M, WebVid-2M } &\gc{39.4}&\gc{68.8}&\gc{78.3}&\gc{2.0}\\

\gc{\textbf{OA-Trans$\ddag$[12F]}} & &  \gc{CLIP-WIT}  & \gc{[5.5M] CC3M, WebVid-2M}  &\gc{40.9}&\gc{70.4}&\gc{80.3}&\gc{2.0}\\
\shline
\multicolumn{8}{c}{\textit{Zero-shot}} \\
HT MIL-NCE~\cite{miech2019howto100m} &CVPR'20& - &[136M] HowTo100M & 7.5 & 21.2 & 29.6 & 38.0     \\
SupportSet~\cite{patrick2020support} &ICLR'21& IG65M, ImageNet &[136M] HowTo100M & 8.7 & 23.0 & 31.1 & 31.0     \\
Frozen \cite{bain2021frozen} &ICCV'21& ImageNet &[2.5M] WebVid-2M  & 14.5 & 29.5 & 64.5 & 21.0 \\
Frozen \cite{bain2021frozen} &ICCV'21& ImageNet & [5.5M] CC3M, WebVid-2M & 18.7 & 39.5 & 51.6 & 10.0 \\
\rowcolor{LightCyan}
Frozen \cite{bain2021frozen} [Our Imp.] &ICCV'21 & ImageNet  & [5.5M] CC3M, WebVid-2M & 21.7 & 45.5& 53.9 & 9.0 \\
CLIP[12F] \cite{radford2021learning} &Arxiv'21&CLIP-WIT&-&28.5&49.7&61.2&5.0\\
\textbf{OA-Trans} && ImageNet& [2.5M] WebVid-2M  & \textbf{18.4} & \textbf{36.5} & \textbf{46.8} & \textbf{10.0} \\
\textbf{OA-Trans} && ImageNet & [5.5M] CC3M, WebVid-2M & \textbf{23.4} & \textbf{47.5} & \textbf{55.6} & \textbf{8.0} \\

\gc{\textbf{OA-Trans}$\ddag$} & & \gc{CLIP-WIT} & \gc{[5.5M] CC3M, WebVid-2M}  &\gc{29.7}&\gc{52.1}&\gc{63.5}&\gc{5.0} \\
\gc{\textbf{OA-Trans}[12F] $\ddag$} & & \gc{CLIP-WIT} & \gc{[5.5M] CC3M, WebVid-2M} &\gc{31.4}&\gc{55.3}&\gc{64.8}&\gc{4.0} \\

\end{tabular}
\caption{\label{tab:msr-vtt-sota}Comparison with state-of-the-art results on MSRVTT for text-to-video retrieval. 
$\ddag$ denotes the model is initialized with weights from CLIP~\cite{radford2021learning}.
\textbf{Vis Enc. Init.:} Datasets that visual encoders' initial weights are trained on.}
\end{table*}

\subsection{Object-aware Transformer}
\label{sec:oa_transformer}

In this section, we present our efficient and simple Object-aware Transformer (OA-Trans) in detail. The pipeline of OA-Trans is shown in Fig.~\ref{fig:3_ppl}.
The distinction from the baseline is the additional masked image stream and object tag stream.
Given an input pair of video and text, we first sample one video clip from this video.
Then we find the central index from this clip and find the closest object frame.
From this object frame, we generate the masked \textbf{\textit{anchor}} object image and object tags.

Instead of using \emph{cls} token, for the masked image we average tokens from non-masked patches and the normalized embedding is represented as $v_l$.
Similarly, the output for the object tag stream is represented as $t_l$.
Then we compute the matching loss $\mathcal{L}_{M}$ and Object-aware Contrastive (OAC) loss  $\mathcal{L}_{OAC}$ from their corresponding output.
Next, we introduce the key components and their design intention of this pipeline as below:

\paragraph{Anchor Object Frame.}
Given a video with arbitrary length, we first uniform sample $L$ (\emph{i.e.}, $L=8$) frames, and an improved Faster RCNN \cite{ren2015faster} \footnote{https://github.com/MILVLG/bottom-up-attention.pytorch} is used to extract $N$ objects offline (probably over-sampled and noisy).
We save these offline objects on disk for reuse.
During training, we select top-$N$ objects with unique object categories.
If the object is too large, we reduce the object size to half of it.
The analysis of the object number is provided in Sec.~\ref{ablation}.

\paragraph{Masking.} 
Given an object frame with $N$ object regions, we first \textit{mask} the region that does not contain objects.
We then divide an \textit{masked} frame into regular non-overlapping patches.
Then we sample a subset of patches that contain the object region and mask the remaining ones to form a regular grid.
In this way, a patch will be either masked out or keep its original pixels.
To prevent overfitting, we drop 20\% objects randomly and shift the \textit{\textbf{anchor}} frame to an adjacent frame in time.
In addition, we crop the central region if the object region is too large.
We simply refer to this as "object-guided masking".
With the proposed Object-Guided Masking, the model is forced to learn to understand the context information and relationships of objects, rather than simply modeling scene bias.

\paragraph{Object-aware Contrastive (OAC) Loss.}
Since our aim is to enhance the fine-grained representation, the straightforward idea is to align the predicted object tags and the local masked image directly.
However, this naive approach will not be able to directly benefit downstream applications because objects are not input to the model for downstream application.
And the loss is quite easy to optimize and may fail into trivial solutions and further damage the learning of global video to text matching.
Based on this observation, we propose a novel OAC loss with cross guidance from object regions to captions and from object tags to video frames.

Specifically, we first use object tags to align with raw video.
Formally,
\begin{equation}
\mathcal{L}_{tag} = -\log \frac{\exp \left(sim\left(v, t_{l} \right) / \tau\right)}{\sum_{i=0}^{K} \exp \left(sim\left(v^i, t_{l} \right) / \tau\right)}
\end{equation}

Although the object tags are from a limited 1600-class dictionary defined by Visual Genome \cite{krishna2017visual}, the tags are usually capable of capturing relevant high-level semantics presented in captions.
For example, \textit{Woman} and \textit{Visitors}, \textit{Tree} and \textit{Garden} in the Fig. \ref{fig:3_ppl}.
Then if we encourage the global visual embedding $v$ not only to align with $t$ but also $t_l$, the model will strengthen the association between different nouns potentially.

Similarly,  we force the model to align the full sentence with a masked object frame.
Formally,
\begin{equation}
\mathcal{L}_{mask} = -\log \frac{\exp \left(sim\left(v_l, t \right) / \tau\right)}{\sum_{i=0}^{K} \exp \left(sim\left(v_l, t^i \right) / \tau\right)}.
\end{equation}
Combining these complementary cross guidance, we define the OAC Loss as:
$\mathcal{L}_{OAC}=\mathcal{L}_{tag}+\mathcal{L}_{mask}$

\paragraph{Overall Training Objective.} The final loss function of OA-Transformer is:
\begin{equation}
    \mathcal{L} = \mathcal{L}_{M} + \lambda \mathcal{L}_{OAC},
\end{equation}
where $\lambda$ is the coefficient that controls the balance between global match loss and OAC loss.

By forcing both video encoder and text encoder to mine object-centric information, our video-text model directly benefits from the high-level semantics captured by object regions and object tags.
As a result, the OA-Trans learns more discriminative representations for downstream video-text tasks.

\subsection{Discussion}
\label{sec:discussion}

\paragraph{Advantages.}
There exist several advantages for the OA-Trans:
\emph{i}. We only use one object image as reference during pretraining and the extra computation cost is limited.
\emph{ii}.
The object knowledge is learned during pretraining, thus reducing the effects of noisy objects on downstream tasks.
\emph{iii}. Our model does not have the need of modifying the architecture of base vision encoder that can be plug-and-play into existing video-language pretraining methods.

\paragraph{More Ways to Incorporate Objects.}
Besides the simple masking operation, we also empirically studied multiple ways to use objects in both vision and language modality inspired by previous works \cite{li2020oscar,kim2021vilt}.
For visual modality, we consider \textit{Pure Offline Features} and \textit{The joint modeling of Offline Feature with Raw-pixel Video}.
All these design details are presented in the supplementary.
We compare all design choices and show our solution is the superior design.

\section{Experiments}

We evaluate our Object-aware Transformer (OA-Trans) on several video-text benchmarks.
Specifically, we consider the following tasks: Video-Text Retrieval (Section~\ref{video-text retrieval}) and Linear Probe Evaluation (Section~\ref{liner probe evaluation}).

\subsection{Pretraining Datasets} 
Since the widely-used dataset, i.e., HowTo100M~\cite{miech2019howto100m}, is heavily noisy and only contains instructional videos. 
In this work, we adopt two clean datasets: (\textbf{\emph{i}}) WebVid2.5M (video-text); and (\textbf{\emph{ii}}) Google Conceptual Captions (image-text) to cover more generalized scenarios.

\noindent\textbf{WebVid2.5M} \cite{bain2021frozen} consists of 2.5M video-text pairs, which is an open domain video captioning dataset. The manually generated captions are well-formed sentences. 

\noindent\textbf{Google Conceptual Captions (CC3M)} is scraped from the web and more than 10\% of CC3M images are in fact thumbnails from videos.
As some images are missing in the web, we get 2.97M images in total.

\subsection{Downstream Datasets}
To verify the effectiveness of learned visual and textual representations, we evaluate OA-Trans on four video-text benchmarks as follows:

\noindent\textbf{MSRVTT} \cite{xu2016msr} contains
10K YouTube videos with $200$K descriptions. 
Following the previous works \cite{Liu19a,bain2021frozen}, we use $9$K videos for training and report results on the $1$K test set. 

\noindent\textbf{DiDeMo} \cite{anne2017localizing} contains $10$K Flickr videos. Each video is annotated with multiple captions, which results in $40$K sentences in total. In the experiments, all captions of a video are regarded as a single description. 

\noindent\textbf{MSVD} \cite{chen2011collecting} contains $20$K YouTube videos annotated with 100K sentences. 
The training set contains 10K videos, and we report results on the validation set with 4.9K videos. 
Since each video is annotated with multiple sentences, we report both \textit{Sentence to Video} and \textit{Multiple Sentences to Video} results to compare with related works.

\noindent\textbf{LSMDC} \cite{Rohrbach_2015_CVPR} contains 120K video-text pairs from $202$ movies.
Following \cite{rohrbach2017movie}, the validation set contains 7K pairs, and evaluation is conducted 1K test set.

\newcolumntype{g}{>{\columncolor{Gray}}c}
\begin{table}
\centering
\footnotesize
\begin{tabular}{@{}lrrrg@{}}
\textbf{Method}  & \textbf{R@1}  & \textbf{R@5}  & \textbf{R@10} & \textbf{MedR} \\ 
\shline
\multicolumn{5}{c}{\textit{Sentence to Video}}\\
Multi. Cues~\cite{mithun2018learning}            & 20.3          & 47.8          & 61.1          & 6.0           \\
CE~\cite{Liu19a}                   & 19.8          & 49.0          & 63.8          & 6.0           \\
Support Set~\cite{patrick2020support} (HowTo PT)   & 28.4          & 60.0          & 72.9          & 4.0           \\
Forzen \cite{bain2021frozen}   & 33.7 & 64.7 & 76.3 & 3.0  \\ 
\textbf{OA-Trans}    & \textbf{39.1} & \textbf{68.4} & \textbf{80.3} & \textbf{2.0}  \\ 
\hline
\multicolumn{5}{c}{\textit{Multiple Sentences to Video}}\\
TeacherText \cite{croitoru2021teachtext} & 25.4&56.9&71.3&4.0 \\
CLIP4CLIP \cite{luo2021clip4clip}   & 46.2 & 76.1 & 84.6 & 2.0  \\ 
\textbf{OA-Trans}   & \textbf{51.4} & \textbf{82.3} & \textbf{88.0} & \textbf{2.0}  \\ 
\end{tabular}
\caption{Text-to-video retrieval results on MSVD~\cite{chen2011collecting}.}

\label{tab:msvd-sota}
\end{table}

\subsection{Setup}
\noindent\textbf{Backbone.} The main components of our method are Visual Encoder and Textual Encoder.
For the Textual Encoder, we adopt Distill Bert \cite{distilbert} as default.
For the Visual Encoder, we adopt Vision Transformer with space-time attention from TimeSformer \cite{bertasius2021space}.
For the Vision Transformer, the 12-layer ViT-B/16 is used as the backbone.
All models trained for 128 epochs.

\noindent\textbf{Technical Detail.} We use the Adam optimizer with weight decay regularization and decay the learning rate with a cosine schedule.
When pretraining on WebVid2.5M, 1 object reference frame and 4 video frames are sampled. 
For CC3M, the video frame number is set to 1 because CC3M is an image-text dataset.
The control weight $\lambda$ is set to 0.5 experimentally.

The whole pretraining takes 5 days on 64 Tesla A100 GPUs.
Unless otherwise specified, all results reported in this paper adopt the best model.
When fine-tuning the pretrained model, only 8 video frames are sampled on all downstream tasks.

\subsection{Video-Text Retrieval}
\label{video-text retrieval}

\noindent\textbf{MSRVTT.} Table \ref{tab:msr-vtt-sota} summarizes the results on \textbf{MSRVTT}.
Besides ClipBERT and Support Set, other methods are pretrained on 136M clip-caption
pairs from HowTo100M.
To ensure a fair comparison, we re-implement the previous SOTA method, Frozen~\cite{bain2021frozen}, with a distributed parallel training schedule.
Under the full fair comparison, OA-Trans outperforms the previous best method Frozen by 2.6\% on R@1.
Surprisingly, only pretrained with open domain 2.5M video-text pairs, our method already outperforms all previous works that are pretrained on 136M clip-caption pairs.

Typically, to evaluate the generalization of models, we also report zero-shot results, i.e., no fine-tuning is conducted.
Our method outperforms previous methods significantly.
The results show that our model has a better generalization ability than others.
To further verify our method can extend to strong visual backbones, we initialize the visual encoder with CLIP's weights~\cite{radford2021learning}. As the results shown in Table \ref{tab:msr-vtt-sota}, our method still improves the performance of CLIP. Thus, our method works well with different initial weights even if the loaded initial weights already have a strong performance. 

\noindent\textbf{MSVD.} Because each video is annotated with multiple captions, previous works are mainly divided into two types:
\emph{i}. Sentence to video: Treat each sentence as the textual query.
\emph{ii}. Multiple sentences to video: Combine multiple sentences of a video as the textual query.
The results are shown in Table \ref{tab:msvd-sota}, in both settings, our method outperforms other methods by 5\% on R@1 at least.

We also show the retrieval results on \textbf{DiDeMo} and \textbf{LSMDC} in Table \ref{tab:didemo-sota} and Table \ref{tab:lsmdc-sota}.
OA-Trans outperforms previous methods on all metrics.

\newcolumntype{g}{>{\columncolor{Gray}}c}
\begin{table}
\centering
\footnotesize
\begin{tabular}{@{}llrrg@{}}
\textbf{Method}& \textbf{R@1}  & \textbf{R@5}  & \textbf{R@10} & \textbf{MedR} \\ 
\shline
S2VT~\cite{venugopalan2014translating}            & 11.9          & 33.6          & -             & 13.0          \\   
FSE~\cite{zhang2018cross}          & 13.9          & 36.0          & -             & 11.0          \\
CE~\cite{Liu19a}  & 16.1          & 41.1          & -             & 8.3           \\
ClipBERT~\cite{lei2021less}          & 20.4          & 44.5          & 56.7          & 7.0           \\
Frozen \cite{bain2021frozen}     & 31.0 & 59.8 & 72.4 & 3.0  \\
\textbf{OA-Trans}      & \textbf{34.8} & \textbf{64.4} & \textbf{75.1} & \textbf{3.0}    \\
\hline
\multicolumn{5}{c}{\textit{Zero-shot}} \\
Frozen \cite{bain2021frozen}  & 21.1 & 46.0 & 56.2 & 7.0 \\
\textbf{OA-Trans}   & \textbf{23.5} & \textbf{50.4} & \textbf{59.8} & \textbf{6.0} \\
\end{tabular}
\caption{Text-to-video retrieval results on DiDeMo. We show both the fine-tune and zero-shot retrieval results .}

\label{tab:didemo-sota}
\end{table}
\newcolumntype{g}{>{\columncolor{Gray}}c}
\begin{table}
\centering
\footnotesize
\begin{tabular}{@{}lrrrg@{}}
\textbf{Method}  & \textbf{R@1}  & \textbf{R@5}  & \textbf{R@10} & \textbf{MedR} \\ 
\shline
JSFusion~\cite{yu2018joint}     & 9.1          & 21.2          & 34.1          & 36.0  \\
MEE~\cite{miech18learning}      & 9.3          & 25.1          & 33.4          & 27.0  \\
CE~\cite{Liu19a}                & 11.2          & 26.9          & 34.8          & 25.3  \\
Forzen \cite{bain2021frozen} & 15.0 & 30.8 & 39.8 & 20.0 \\
MMT (HowTo PT)~\cite{gabeur2020multi}                & 12.9          & 29.2          & 38.8          & 19.3  \\
\textbf{OA-Trans}                   & \textbf{18.2} & \textbf{34.3} & \textbf{43.7} & \textbf{18.5}  \\
\end{tabular}
\caption{Text-to-video retrieval results on LSMDC.}
\label{tab:lsmdc-sota}
\end{table}

\begin{figure*}[t]
  \centering
   \includegraphics[width=.9\linewidth]{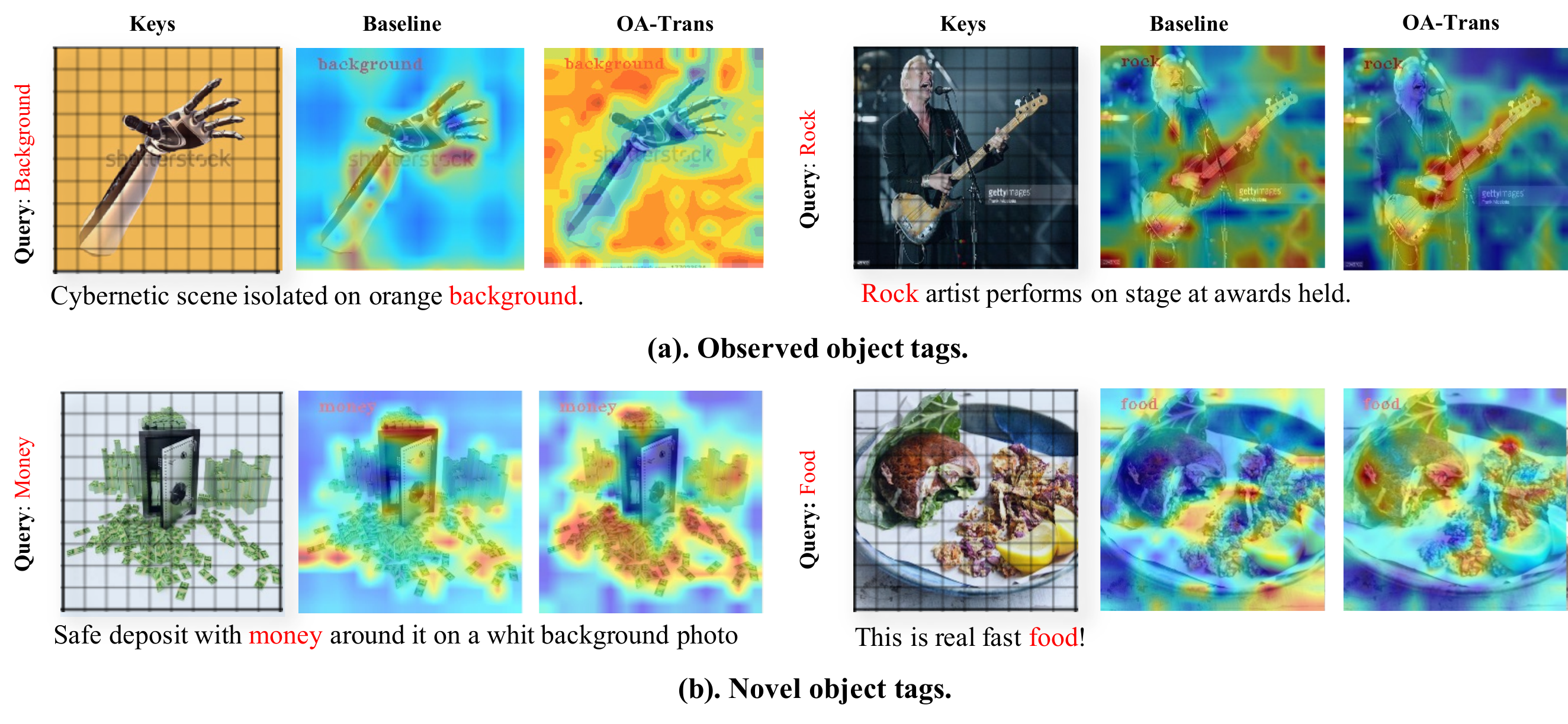}
   \caption{
   \textbf{Cross-modality attention regions visualization.}
   The specific text token as query and the patch-level tokens as keys.
   In the upper part, ``people'' and ``rock'' are in the predefined object vocabulary. In the bottom part, ``food'' and ``money'' are not covered by the predefined object vocabulary.
   }
   \label{fig:4_att_vis}
\end{figure*}

\subsection{Linear Probe Evaluation}
\label{liner probe evaluation}
The linear probe is an important measurement to evaluate the quality of representations learned in large-scale image-text pretraining \cite{radford2021learning} and image self-supervised pretraining \cite{he2020momentum}.
However, this technique is never explored in video-text pretraining and most related works still focus on fine-tuning the overall model.

The fine-tune strategy brings two problems:
\emph{i}. The hyper-parameter spaces for various downstream datasets are very large.
It's very difficult to provide fair comparisons among different pretrain methods.
\emph{ii}. Fine-tuning adjusts the overall model and adapts representations to a specific dataset, it may hide the failures that a model does not learn general and robust representations.

Following CLIP \cite{radford2021learning}, in this work we fit a linear classifier on representations extracted from the pretrained model and measure its performance on various downstream datasets. 
We implement Frozen and CLIP by ourselves.
Since CLIP is an image-text pretrain method, we sample 8 frames of each video and average the image-level feature to represent a video.
The results are shown in Table \ref{tab:linear_prob}.
We also show the results of OA-Trans initialized with CLIP-pretrained weights.
It can be seen that OA-Trans generalizes well to these datasets.
We hope this experiment will inspire the community to focus more on this task.

\begin{table}
\centering
\footnotesize
\begin{tabular}{@{}llrrrr@{}}
\textbf{Method} &\textbf{VE Init.}  & \textbf{MSR}  & \textbf{MSVD}  & \textbf{DiDeMo} & \textbf{LSMDC}\\ 
\shline
Frozen \cite{bain2021frozen}     &ImageNet& 27.2        & 30.3         & 26.6          & 13.2  \\
\textbf{OA-Trans} & ImageNet   & \textbf{31.3}         & \textbf{34.1}          & \textbf{30.4}          & \textbf{18.1}  \\
Clip\cite{radford2021learning} &CLIP-WIT & 30.5& 34.5& 29.8&16.8\\
\gc{\textbf{OA-Trans $\ddag$}}             & \gc{\textbf{CLIP-WIT}}& \gc{\textbf{33.2}}         & \gc{\textbf{36.9}}          & \gc{\textbf{34.8}}          & \gc{\textbf{21.5}}  \\
\end{tabular}
\caption{The linear probe evaluation of three video-text retrieval datasets. $\ddag$ means we use CLIP weight for visual encoder initialization.
We report R@1 result and VE Init is short for Visual Encode Initialization.}
\label{tab:linear_prob}
\end{table}

\subsection{Qualitative Visualization}


\paragraph{Attention Region Visualization.}
To provide insight into the inner representation of OA-Trans, we provide further visualization.
Specifically, we visualize the attention map between captions and visual patches, where a text token is regarded as the query and attention weights on all spatial tokens are visualized.
We use the output of the first Transformer layer for visualization.
To analyze if OA-Trans only helps the modeling of nouns that are included in object tag dictionary. 
We select nouns from both the object tag dictionary and other novel object tags that are not included in the object tag dictionary.

The visualization of the attention weights allocated to each patch is shown in Fig. \ref{fig:4_att_vis} and we make the following observations:
\emph{i}. For the complex scenarios like ``awards held" in the up-right of Fig. \ref{fig:4_att_vis}, 
OA-Trans focuses on rock devices more accurately while baseline looks at irrelevant corners.
\emph{ii}.
Interestingly, even ``money" and ``food" are not included in the object tag dictionary,
OA-Trans still focuses on the corresponding regions accurately.
\textit{This experiment demonstrates the introduction of object tags and regions improves the overall representation ability rather than fits an implicit bias over object tags}.

\begin{table}
\footnotesize
\centering
\begin{tabular}{@{}ll|rrr|rrr@{}}
\textbf{$\mathcal{L}_{tag}$} & \textbf{$\mathcal{L}_{mask}$} 
& \multicolumn{3}{c|}{\textbf{T2V}}
& \multicolumn{3}{c}{\textbf{V2T}}\\
&&R@1&R@5&R@10&R@1&R@5&R@10 \\ 

\shline
& & 14.5 &31.6&40.8&14.8&29.7&40.6   \\
\textcolor{darkergreen}{\checkmark}      & & 17.4 & 33.2 &45.7 & 18.1 &33.6 & 42.7    \\
 & \textcolor{darkergreen}{\checkmark} & 15.9 & 33.2 &43.3 & 15.4 &30.9 & 40.8 \\
\textcolor{darkergreen}{\checkmark}     & \textcolor{darkergreen}{\checkmark} & 18.4 &36.2&47.8&17.5&33.0&46.4    \\ 

\end{tabular}
\caption{The ablation of object category and object region on MSRVTT.
$\mathcal{L}_{tag}$ means object tags to video match loss and $\mathcal{L}_{mask}$ means object mask image to text match loss.
}
\label{tab:abl_comp}
\end{table}

\subsection{Ablation Studies}
\label{ablation}
In this section, we conduct ablation studies and analyze the different choices for utilizing object information.
We pretrain our OA-Trans on WebVid2.5M and conduct an evaluation on zero-shot MSRVTT retrieval.

\paragraph{Effectiveness of Each Component.}
In this section, we explore the effect of object region and object tag.
The results are given in Table \ref{tab:abl_comp}.
When using object tag, our method achieves 1.4\% R@1 gain compared to the baseline in text-to-video retrieval.
We also find object tags contribute more to the retrieval ability.
The combination of object tag and object region leads to the best result.

\begin{figure}
  \centering
   \includegraphics[width=.9\linewidth]{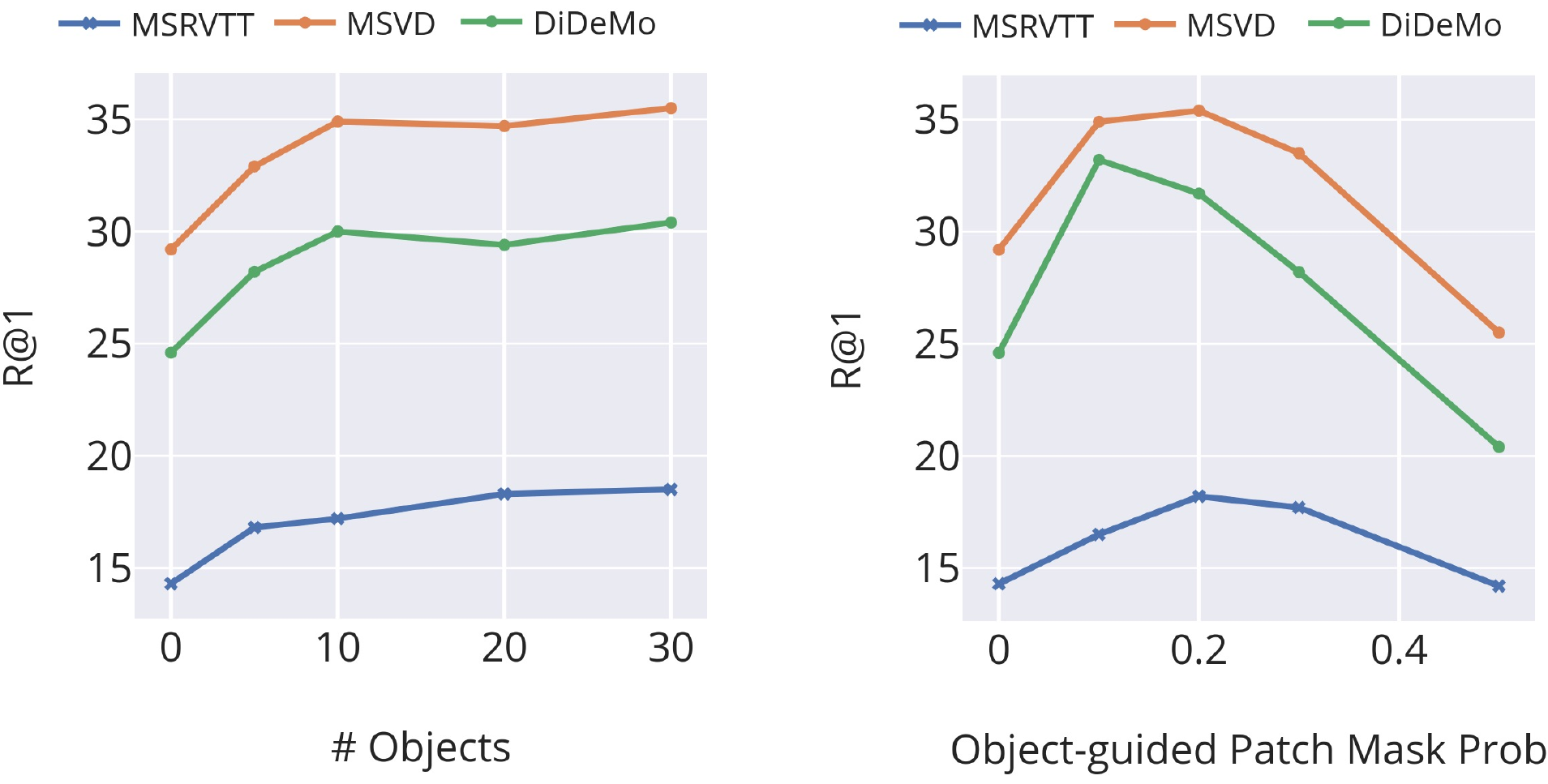}

   \caption{
   Left: The number of object and the corresponding Retrieval top-1 result.
   Right: The object-guided mask probability and the corresponding R@1 result.
   }
   \label{fig:4_object_abl}
\end{figure}

\paragraph{Number of Objects.}
In the left of Fig.~\ref{fig:4_object_abl}, we compare the results of different OA-Trans by varying the number of objects.
We find that more objects lead to better performance in general.
When the number of objects is larger than 10, the performance remains consistent. Thus, the number of objects is set to 10 as default.

We also explore the impact of the mask patch probability in the right of Fig. \ref{fig:4_object_abl}.
For this experiment, we take mask probability from 0 to 0.5 for comparison.
We can see that the accuracy grows firstly as the probability increases for all three datasets.
But when the probability is larger than 0.2, all results drop significantly.
The large mask probability will drop too many regions and the semantics may change.

\paragraph{Strategy of Object Tag Utilization.}
In this section, we investigate the different ways to utilize the object tag.
We study three variations:
\emph{i}. \textbf{Padding:} Pad object tags to the original caption as in Oscar \cite{li2020oscar}.
\emph{ii}. \textbf{Two Stream:} 
Use two-stream input. 
One stream is the original caption, the other stream is the object tags.
\emph{iii}. \textbf{Two Stream + Padding:} Use two-stream input. 
One stream is the original caption, the other stream is the original caption with padding object tags.
Notice all the strategies are designed for pretraining.
During testing, we use normal video-text retrieval settings to show the generalization of our method.

\begin{table}
\footnotesize
\centering
\begin{tabular}{@{}l|rrr|rrr@{}}

\textbf{Method}& \multicolumn{3}{c|}{\textbf{T2V}}& \multicolumn{3}{c}{\textbf{V2T}}\\
&R@1&R@5&R@10&R@1&R@5&R@10 \\ 

\shline
Baseline& 14.5 &31.6&40.8&14.8&29.7&40.6   \\
Padding\cite{li2020oscar}& 15.5 &33.2&43.4&15.7&30.5&41.2   \\
Two Stream$\dag$& 17.5 &35.9&47.8&17.5&35.7&46.5   \\
Two Stream& 17.7 &35.5&48.1&18.2&34.7&45.6   \\

\end{tabular}
\caption{\textbf{The variations of utilizing object categories.} Two Stream$\dag$ means Two Stream + Padding.
}
\label{tab:abl_categories}
\end{table}

The results are shown in Table \ref{tab:abl_categories}.
We find padding operation leads to around 1\% improvement on both text-to-video and video-to-text retrieval settings.
The reason behind this phenomenon is that the padding operation performs like an augmentation to the text.
When introducing a two-stream pipeline, we find the R@1 for both text-to-video and video-to-text tasks is improved by around 3\%.
In such a form, the model is asked not only to align a video with its original caption but also the padding of detailed objects. Thus, object information that is not mentioned in the caption is also preserved in the visual representation. Such visual representations could help the pretrained model generalize well to more scenarios.
In this work, we adopt the Two Stream strategy as default.

\paragraph{Alternative Inputs in Visual Stream.}

In this section, we give a comparison between different visual inputs to see which one helps to capture better representations in our OA-Trans.
Specifically, we keep other components unchanged and then we compare three visual inputs as follows:
\emph{i}. \textbf{Raw Video Input}: 
Only input the original video.
\emph{ii}. \textbf{Only Masked Input}: 
We remove the original raw video stream and only input masked anchor image.
\emph{iii}. \textbf{Joint Input}:
Input the masked anchor image and the raw video stream. 

\newcolumntype{g}{>{\columncolor{Gray}}c}
\begin{table}
\footnotesize
\centering
\begin{tabular}{@{}lrrrg@{}}

\textbf{Method} &R@1&R@5&R@10&MedR \\
\shline
Mask Only& 16.4 &35.5&45.8 &11.0  \\
Raw Video Only& 15.9 &33.2&43.3 &12.0   \\
Joint Input& 18.5 &37.2&49.8 &10.0   \\
\end{tabular}
\caption{
\textbf{The comparison of alternative inputs in visual stream}.
We report the zero-shot retrieval result.
}
\label{tab:abl_pure_object}
\end{table}

The results are reported in Table \ref{tab:abl_pure_object}.
Interestingly, we find the Mask Only input already suppresses Raw Video Only around 2.5\% over R@10 metric.
This demonstrates the importance of object-centric modeling in video-text matching.
Compared with single-stream input, the joint input leads to the best result. 
This Phenomenon indicates that these two streams provide complementary information and the model can benefit from object-region guided local alignment.

\section{Conclusion}
Current dual-encoder networks in video-language pretraining lack the learning of fine-grained semantic alignment.
Objects can provide a strong complement for this problem, but their modeling is very challenging for machine vision especially in video.
The OA-Trans we present here makes use of a simple object bounding box and object tags information to generate a contextualized representation of the entire scene.
We note that such integration is particularly natural in cross-modality transformer models, where an object region has the same role in the architecture as the uniformly-spaced patch tokens.

In our current implementation, we use an externally provided offline object detector.
However, it will be interesting to replace the offline bounding boxes with boxes that the model generates itself without strong supervision.
An additional interesting extension is to cluster visual similar regions in an video in a self-supervised fashion,
where the task is to align the clustered video with text.
We leave these challenges to future work.

\section*{Acknowledgement}
This project is supported by the National Research Foundation, Singapore under its NRFF award NRF-NRFF13-2021-0008.

{\small
\bibliographystyle{ieee_fullname}
\bibliography{main}
}

\end{document}